\documentclass[10pt,journal,compsoc]{IEEEtran}
\newcommand{\MyTitle}{Premonition Net, A Multi-Timeline Transformer Network Architecture Towards Strawberry Tabletop Yield Forecasting}
\newcommand{\MyAuthors}{George Onoufriou, Marc Hanheide, and Georgios Leontidis}
\newcommand{\MyKeyWords}{Deep Learning, Strawberry Yield Forecasting, Time Series, Transformers}

\usepackage[style=ieee,backend=biber]{biblatex} 
\usepackage[pdfauthor={\MyAuthors{}},
            pdftitle={\MyTitle{}},
            pdfkeywords={\MyKeyWords{}},
            pdfproducer={Latex with hyperref},
            pdfcreator={pdflatex}]{hyperref}
\ifCLASSINFOpdf
   \usepackage[pdftex]{graphicx}
    \graphicspath{ {./content/img/} }
\else
\fi
\usepackage{tabularx}
\newenvironment{conditions*}
  {\par\vspace{\abovedisplayskip}\noindent
  \tabularx{\columnwidth}{>{$}l<{$} @{${}={}$} >{\raggedright\arraybackslash}X}}
  {\endtabularx\par\vspace{\belowdisplayskip}}

\usepackage{amssymb}
\usepackage{mathtools}

\usepackage[noend]{algpseudocode}
\usepackage{algorithm}

\usepackage{soul}
\usepackage{tikz}
\usetikzlibrary{calc}

\makeatletter
\newif\if@anon

\ifx\blind\undefined
    \@anonfalse
\else
    \@anontrue
\fi

\if@anon
  \newcommand{\highlight@DoHighlight}{
    \fill [outer sep = -15pt, inner sep = 0pt, color=black]
          ($(begin highlight)+(0,8pt)$) rectangle ($(end highlight)+(0,-3pt)$) ;
  }

  \newcommand{\highlight@BeginHighlight}{
    \coordinate (begin highlight) at (0,0) ;
  }

  \newcommand{\highlight@EndHighlight}{
    \coordinate (end highlight) at (0,0) ;
  }

  \newdimen\highlight@previous
  \newdimen\highlight@current
  \newlength{\item@width}

  \DeclareRobustCommand*\anon{%
    \SOUL@setup
    \def\SOUL@preamble{%
      \begin{tikzpicture}[overlay, remember picture]
        \highlight@BeginHighlight
        \highlight@EndHighlight
      \end{tikzpicture}%
    }%
    \def\SOUL@postamble{%
      \begin{tikzpicture}[overlay, remember picture]
        \highlight@EndHighlight
        \highlight@DoHighlight
      \end{tikzpicture}%
    }%
    \def\SOUL@everyhyphen{%
      \discretionary{%
        \SOUL@setkern\SOUL@hyphkern
        \SOUL@sethyphenchar
        \tikz[overlay, remember picture] \highlight@EndHighlight ;%
      }{%
      }{%
        \SOUL@setkern\SOUL@charkern
      }%
    }%
    \def\SOUL@everyexhyphen##1{%
      \SOUL@setkern\SOUL@hyphkern
      \settowidth{\item@width}{##1}%
      \makebox[\item@width]{}%
      \discretionary{%
        \tikz[overlay, remember picture] \highlight@EndHighlight ;%
      }{%
      }{%
        \SOUL@setkern\SOUL@charkern
      }%
    }%
    \def\SOUL@everysyllable{%
      \begin{tikzpicture}[overlay, remember picture]
        \path let \p0 = (begin highlight), \p1 = (0,0) in \pgfextra
          \global\highlight@previous=\y0
          \global\highlight@current =\y1
        \endpgfextra (0,0) ;
        \ifdim\highlight@current < \highlight@previous
          \highlight@DoHighlight
          \highlight@BeginHighlight
        \fi
      \end{tikzpicture}%
      \settowidth{\item@width}{\the\SOUL@syllable}%
      \makebox[\item@width]{}%
      \tikz[overlay, remember picture] \highlight@EndHighlight ;%
    }%
    \SOUL@
  }
\else
  \newcommand{\anon}[1]{#1}
\fi
\makeatother

\begin{document}
%
\title{\MyTitle{}}
%
%
%
%

\author{\MyAuthors{}
\IEEEcompsocitemizethanks{\IEEEcompsocthanksitem G. Onoufriou and M. Hanheide are both with the University of Lincoln.\protect\\
E-mail: GOnoufriou at lincoln dot ac dot uk
\IEEEcompsocthanksitem Georgios Leontidis is with with the University of Aberdeen.}
\thanks{Manuscript received \today; revised never.}}

%
%

\markboth{Journal of \LaTeX\ Class Files,~Vol.~14, No.~8, August~2015}%
{Shell \MakeLowercase{\textit{et al.}}: Bare Demo of IEEEtran.cls for Computer Society Journals}
%



\IEEEtitleabstractindextext{%
\begin{abstract}
Yield forecasting is a critical first step necessary for yield optimisation, with important consequences for the broader food supply chain, procurement, price-negotiation, logistics, and supply. However yield forecasting is notoriously difficult, and oft-inaccurate. Premonition Net is a multi-timeline, time sequence ingesting approach towards processing the past, the present, and premonitions of the future. We show how this structure combined with transformers attains critical yield forecasting proficiency towards improving food security, lowering prices, and reducing waste. We find data availability to be a continued difficulty however using our premonition network and our own collected data we attain yield forecasts 3 weeks ahead with a a testing set RMSE loss of ~0.08 across our latest season.
\end{abstract}

}

\maketitle

\IEEEdisplaynontitleabstractindextext

%
\IEEEpeerreviewmaketitle

\IEEEraisesectionheading{\section{Introduction}\label{sec:introduction}}

\begin{figure}[t!]
  \centering
  \includegraphics[width=1\columnwidth]{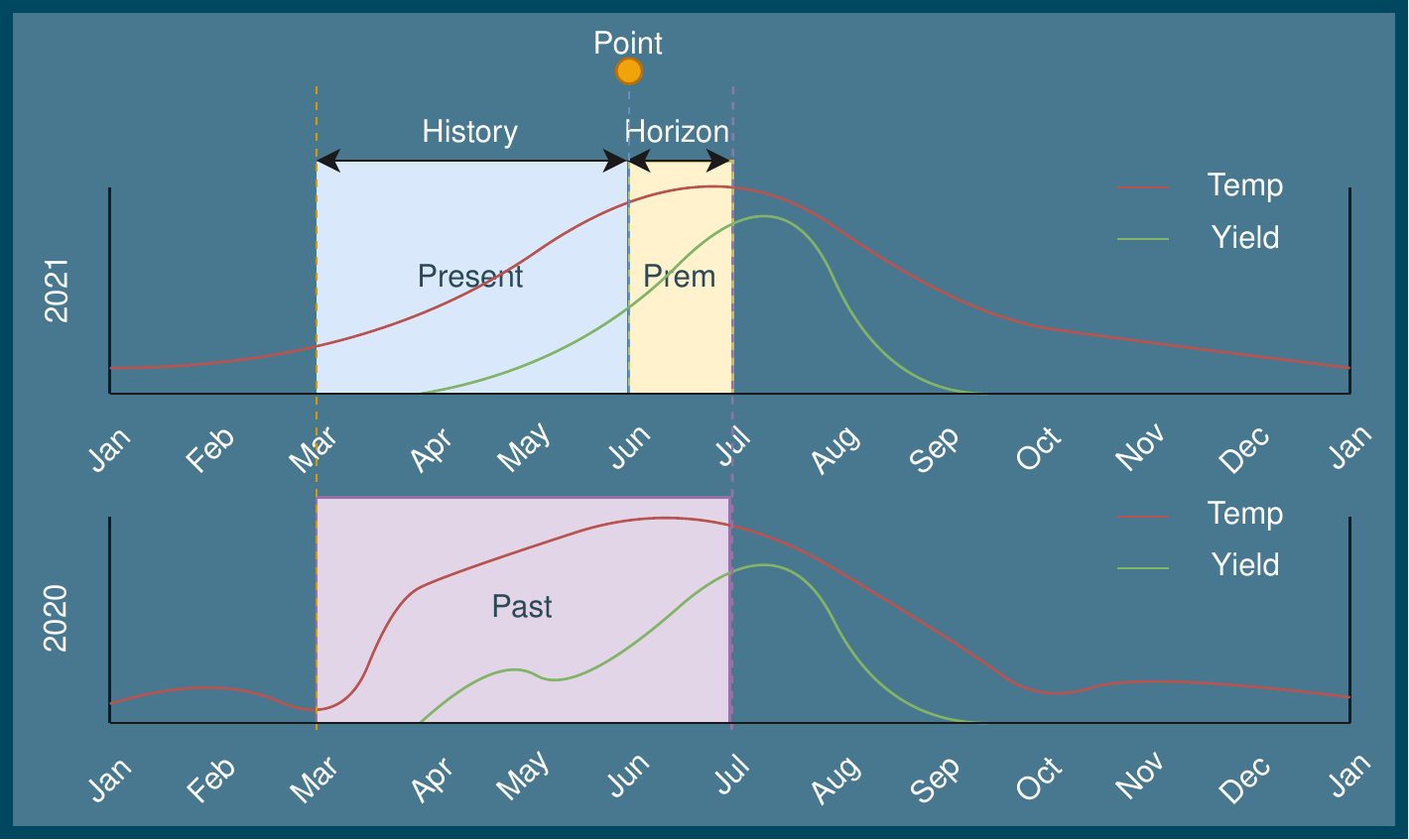}
  \caption{Past (purple-pink), present (blue) and premonition (yellow) timelines/ windows overlayed on a depiction / rough reference of strawberry yields through the years of 2020 and 2021 along with temperature. Depicting the point of prediction relative to (at the seam of) horizon and history.}
  \label{fig:ppp}
\end{figure}

%
%
%
%
\IEEEPARstart{P}{recise} and accurate yield forecasting is a key component in Fresh Produce (FP) Supply Chain Management (FSCM), since it plays a critical role in price negotiations, logistics, and scheduling. In particular accurate yield estimates are required a minimum of 3 weeks ahead (in the strawberry domain) which we call the horizon (Figure \ref{fig:ppp}), so that adequate time can be given to bidding, labour timetabling, logistics, and procurement. However, forecasting FP is incredibly difficult especially over a 3-week horizon where any number of variabilities can exist such as environmental fluctuations. Often the quantities of fresh produce we seek to deal with make it impractical to expect climate-controlled greenhouse conditions, meaning there is an element of weather forecasting that is required however we do not expressly aim to forecast weather in this work as this is a separate and highly complex problem of its own. Instead, we show how good yield forecasting can be and improve upon current practices while allowing for future works to delve specifically into weather forecasting.

Yield forecasting is difficult in particular due to the in-availability of data with which to forecast, this data being mostly non-existent, or incredibly difficult to attain. We believe the reasons why the data is unavailable is because of the difficulty of data collection, the perceived sensitivity with which this data is held, and the lack of clear benefits to the digital collection of such data. We also see resistance to the positive dynamic impetus of modernisation requiring a departure from growers' previous fixed practices.

FP optimisation is of global strategic importance since horticulture and agriculture are some of the biggest producers of greenhouse gasses, such that there can be a significant benefit to optimising production or minimising waste. In the UK our government has committed to reducing greenhouse gasses to net 0 by 2050, and agriculture has been expressly named as a key contributor of greenhouse gasses in the United Nations Climate Change Conference 2021 (COP21). Inaccurate forecasting or more specifically under/ over estimation leads to food waste and destruction costs or importing of FP from abroad. Assuming the cause of this discrepancy/ variability is adverse weather conditions, then those same weather conditions will have affected geographically approximate growing sites. In the UK climate discrepancies usually mean fruit must be imported from abroad, given our size, to meet any given procurement contract, as all the neighbouring growing sites will have suffered the same adverse environmental conditions and thus under-production.

Other works (outlined with more detail in Section \ref{sec:literature}) have sought to solve the lack of data availability in agriculture using satellite/ remote-sensing data, using various machine learning, statistical, and some deep learning techniques. In this paper we show how we can collect data at some scale but with local/ high granularity, including fruit images, weather conditions, and irrigation data locally. Here we shall focus specifically on strawberry yields of strawberry tabletop and how we can predict them. We exemplify this approach at our \anon{Riseholme} strawberry tabletop/ polytunnel growing site and employ this data to create accurate forecasts with this 3-week horizon/ window to meet the needs of the bidding and procurement process. We do all this in collaboration with \anon{Berry Gardens Growers (BGG)}, one of the UK's largest soft, and stone fruit producers, and with their direction on industry standards to keep as close to the typical expectations as reasonably possible. We also have fortnightly visits by agronomists to ensure we are growing the strawberries satisfactorily.

We use this data in various neural network architectures in Section \ref{sec:methodology} and evaluate their performance in Section \ref{sec:discussion}, since the literature would suggest that deep learning approaches are the most performant even for FP. Of these new architectures, we showcase our Premonition Network which seeks to improve upon current tabular/ sequence prediction approaches using all three forms of context, the past, the present, and the premonition of the future. We use the past to learn the overarching distribution, we use the present to set some scale and granularity, and we use the premonition for variability from the standard distribution.

\section{Literature}\label{sec:literature}
There are relatively few works in strawberry yield prediction using deep learning, instead the majority focus on statistical machine learning, and almost none that refer to privacy considerations \autocite{hopf2022development, VANDERVELDE2019203, bouras2021cereal, paudel2021machine, zhu2022deep, bali2022emerging, jafari2020, gastli2021, maskey2019}. However, several papers have stressed that a lack of data availability (\autocite{pearson2019distributed, durrant2021might, durrant2022role}), or more specifically a high expense of acquisition which significantly hinders the smooth application of state-fo-the-art neural networks towards the creation of powerful forecasting models \autocite{nassar22, jafari2020, gastli2021, chen2019, maskey2019}. Many of the aforementioned papers largely choose to tackle this lack of data by using satellite imagery although in some cases they use the California strawberry commission data paired with the California strawberry commission irrigation management information system (CIMIS). Unfortunately the data mentioned in these papers is behind multiple walls, and the CIMIS data is currently unavailable from the original source, so while we were able to find an excerpt of the CIMIS data elsewhere we were unable to find the full dataset making it very difficult to compare to.

Many different proposals for methods of predicting / forecasting yield (generically) exist, some using classical machine learning (e.g. \autocite{paudel2021machine}) others such as those by Nassar \autocite{nassar22} use neural networks in their specific case a mixture of CNN, LSTMs, GRUs and some attention heads. However all emphasise the need for better forecasting systems as demand increases and supply decreases due to global factors such as (but not limited to) COVID-19 and the Russia-Ukraine war. Current yield forecasting methods are highly archaic, often times they can be as simple as forecasting the average of the last few years' yields, or simple linear models based on heat hours. \autocite{paudel2021machine} One such example is the European Commission's MARS crop forecasting system (MCYFS) which has purportedly seen no improvement in its forecasting performance since 2006 and uses no machine learning. \autocite{paudel2021machine} Lastly the work by Paudel \autocite{paudel2021machine} shows that machine learning can already at the very least match (at the start of the season) or beat existing large-scale traditional crop yield forecasting systems such as the aforementioned MCYFS system.

The MCYFS system from 2006 to 2015 has a median MAE of 0.379, 0.368, 0.570 in soft wheat durum wheat and grain maize \autocite{VANDERVELDE2019203}. The most performant forecasts for this system appear to be sunflower yields at 0.162 MAE. However the assessment carried out by van der Velde does not state over what period these yield predictions are made specifically whether that be a few weeks, days or months ahead making this also a difficult comparison to make. It is also apparent that forecasting is becoming increasingly difficult with the higher degree of variability in climate conditions as the performance of this largely static forecasting system seems to be in slow decline \autocite{VANDERVELDE2019203}.

As more modern dynamic techniques are still only just beginning to be used in literature towards strawberry tabletop forecasting we look towards the application of these much more modern techniques, in particular deep learning / neural networks. However, as previously stated data is incredibly difficult to attain in this domain. Nassar \autocite{nassar22} appears to show how the compound deep learning models outperform standalone deep learning models and traditional machine learning models. Nevertheless, as with much work in this space, it is difficult to garner any concrete comparable statistics. From one of their diagrams (14) we believe we can see their most performant model to produce an MAE loss of roughly 0.14 or 14\% MAPE. They call this model Attention-ConvLSTM2D. While we do not have access to the same data as they have, we have seen even simple GRU models attain similar performance in our strawberry tabletop. However, we believe we can improve this performance on our own data by means of attention as their paper would also suggest, but instead of standalone attention heads we intend to use a much more complex and performant transformer model.

For the sake of completion, we also take a look at an adjacent work (strawberry counting including flowers) by Yang Chen \autocite{chen2019} which seeks to count strawberries. Clearly, the number of flowers currently available on the strawberries will be directly linked to the outcome of strawberry fruits since it is these pollinated flowers that will attempt to become fruits. Chen uses their own self-collected dataset in Florida using drones or UAVs to capture images of the fruiting strawberries. Using these drones and Faster R-CNN based on ResNet-50 \autocite{rcnn, resnet}.

Transformers as proposed by Vaswani et al \autocite{vaswani2017attention} are state-of-the-art neural network components for sequence-to-sequence problems. Strawberry yield prediction is such a problem thus we are keen to implement and use them in this scenario, having used other methods to varying degrees of success in the past \autocite{ukras2020Pipeline, FHEoNNG}. We also note that in contrast to our previous techniques transformers and their attention heads can help focus the neural network into parts of the data that are most important thus reducing the need for quite as much data compared to equivalently complex neural networks.

In short yield forecasting is essential for improving on food security, and sustainable development \autocite{zhu2022deep}. Yield estimation is difficult due to a lack of data availability and thus a lack of research using modern data-hungry techniques in this domain \autocite{nassar22, jafari2020, gastli2021, chen2019, maskey2019}. Most attempt to solve this data shortfall by using remote sensing, or by using a select few difficult-to-attain datasets like the california commissions data \autocite{zhu2022deep, jafari2020, nassar22}. Few works have applied modern deep learning / neural networks successfully to agriculture, and especially strawberries, the majority use either old neural network forms or don't use neural networks at all.

\section{Methodology}\label{sec:methodology}
\begin{figure*}[h]
  \centering
  \includegraphics[width=1\textwidth]{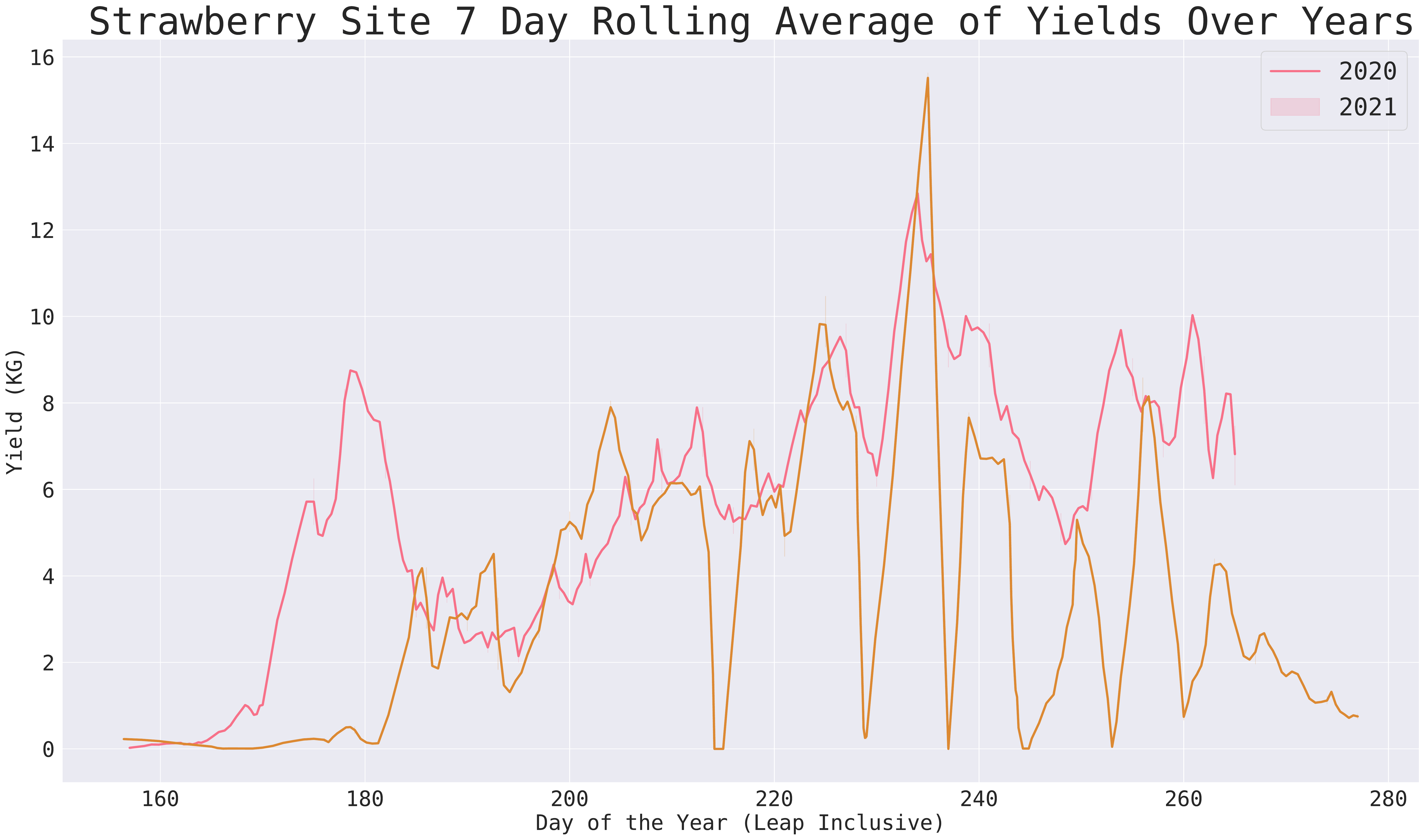}
  \caption{Seven day rolling average line-plot of the strawberry yields of both the 2020 and 2021 seasons.}
  \label{fig:7davg}
\end{figure*}



\begin{figure}[h]
  \centering
  \includegraphics[width=1\columnwidth]{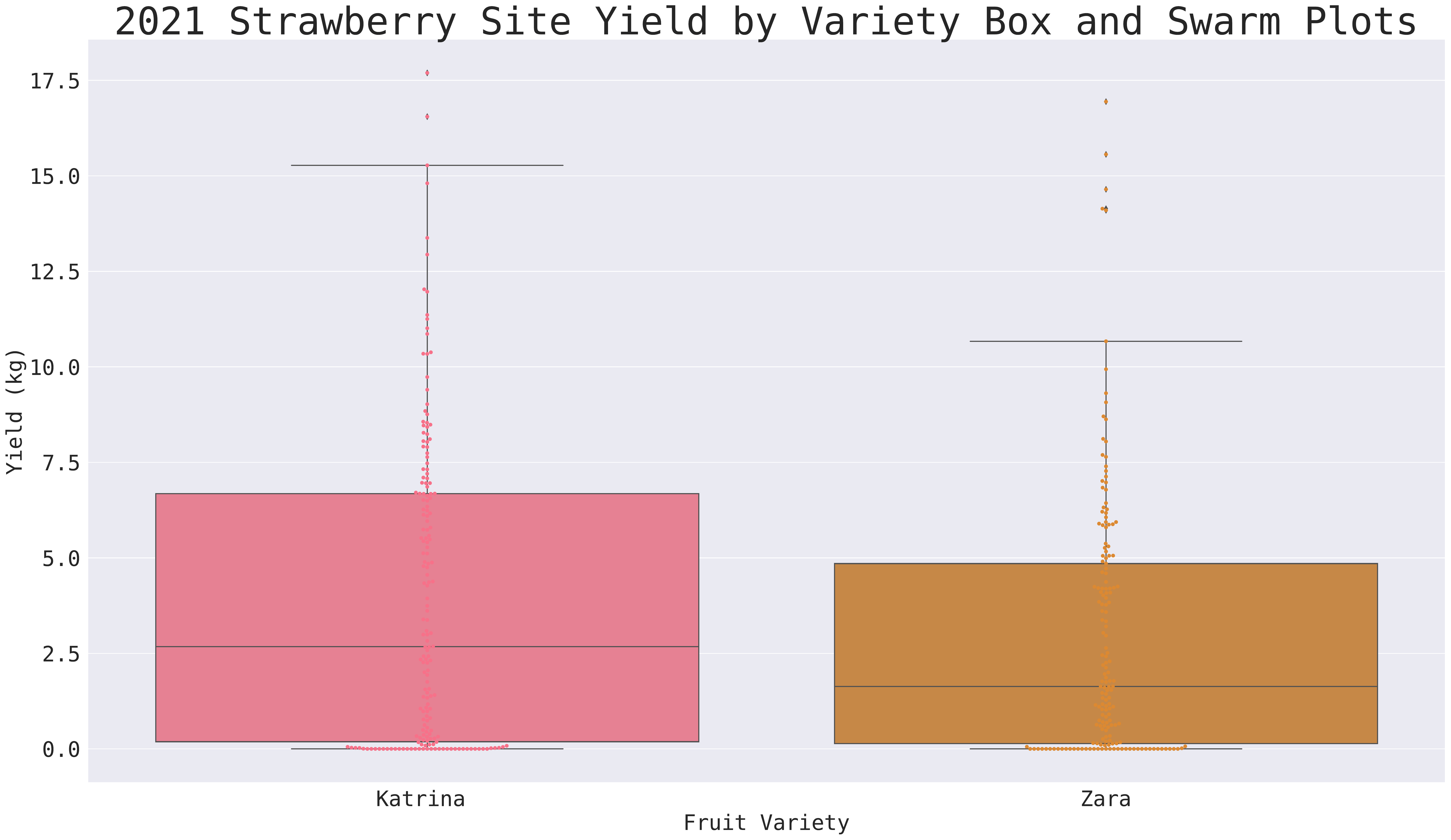}
  \caption{Yield performance of the Katerina and Zara strawberry varieties over the 2021 growing season.}
  \label{fig:boxv}
\end{figure}



We have collected 3 years of strawberry tabletop data at our \anon{Riseholme} campus.
This data comprises 2 polytunnels, each with 5 rows of strawberry tabletop, and each tabletop being 20 meters long. 
Thus in total, we had  200 meters of strawberry tabletop over any single season. Over these rows we had two different June bearing varieties at any one time from \anon{Driscoll's Zara}, \anon{Katerina}, and \anon{Malling Centenary}. The data capture devices we employed for this strawberry tabletop was:
\begin{itemize}
    \item Irrigation data from the tabletop irrigation system. This includes features describing the nutrients, moisture levels, soil temperature, input irrigation, and irrigation runoff. With a sample rate of 1 sample per 2 minutes.
    \item Environmental data from a central weathervane which collected information about: Temperature, humidity, wind direction, wind speed, solar radiance, and precipitation. With a sample rate of 1 sample per 15 minutes.
    \item Yield weight and quality data from our strawberry picking team. With a sample rate of 2 full picks per row per week.
\end{itemize}

\subsection{Data Wrangling}

One of the biggest challenges when working with any time-series dataset is to ensure synchronicity. Since all 3 data sources are sampled at different sometimes overlapping intervals it was necessary to re-sample the datasets to achieve synchronisation. We opted to synchronise over the 15 minutes intervals to match the weathervane data. We later downsampled the synchronised data to a much more manageable 4-hour interval when fed into our MTT.

One of the other challenges when working with any data is missing or unrepresentative samples. Unfortunately in real-world scenarios we always expect to capture some missing or inaccurate data, especially when humans are necessarily involved in the process. We chose to use a forward-fill strategy whereby any missing values are filled with the last known values. The only features not forward-filled are ones that are sampled too infrequently to be able to reasonably forward-fill them. This means any missing values in yields for instance (which are collected bi-weekly) are removed as we cannot reasonably infer them from neighbouring values. 

Now that we have a regular dataset with no missing values we can begin example extraction as per Figure \ref{fig:ppp}. We create hopping windows that end on/ are aligned to observed yield outcomes in the current/ predicted-for year. The window lengths we chose are 21 days for the premonition, 12 weeks for the present and the cumulative period for both combined in the previous year as the past. This way we have information on adverse weather forecasts, current strawberry performance and performance of strawberries at the same site last year. We then create time sequences using expected date ranges. the historic data and when we have specific outcomes for fruit yields. This meant we roughly formed 2 examples for every week in the growing season. We then further split this data by row into training (2,3,4,6,7,8,10), and testing (1,5,9) sets, while further subdividing the training set into training and validation using k-fold cross validation where $k=B_t$ with a batch size of $B_s=32$ which resulted in $B_t=10$ batches. We held out the two final shuffled batches as a per-epoch validation set.
We split in this manner to ensure there is no overlap between training and testing sequences, and it enables us to have a full multi-year view since there are not enough years of data with which to hold out.

Finally we normalised our dataset feature-wise using a basic linear transformation Equation \ref{eq:lin_norm}.
\begin{equation}\label{eq:lin_norm}
x_i^{\prime<t>} = (b - a)\frac{x_i^{<t>} - \textit{min}(x_i)}{\textit{max}(x_i)-\textit{min}(x_i)}+a
\end{equation}
 Where the desired normalised feature value for $x_i$ at timestep $t$ post normalisation $x_i^{\prime<t>}$ is in $[a, b]$. We chose our range to be $[-1, 1]$. We inverted our results to real values using the inversion Equation \ref{eq:lin_norm_inv}.
\begin{equation}\label{eq:lin_norm_inv}
 x_i^{<t>} = (x_i^{\prime<t>} - a)\frac{(\textit{max}(x_i)-\textit{min}(x_i))}{b - a}+\textit{min}(x)
\end{equation}

\subsection{Architecture}
As can be seen in Figure \ref{fig:mtt}, our MTT consists of 3 differently parameterised transformers merged together using a dense layer. Thus our architecture is comprised of 3 encoders, 3 decoders and a dense layer.

\subsubsection{Encoder and Decoder}
As is standard for transformer networks it is necessary to decide upon some form of positional encoding \autocite{vaswani2017attention}. In our case we use a standard fixed positional encoding where even positions are encoded using Equation \ref{eq:pos_encode_even} and odd positions are encoded using Equation \ref{eq:pos_encode_odd}.
\begin{equation}\label{eq:pos_encode_even}
PE_{pos,2i}=\textit{sin}(\frac{pos}{10000^{2i/D}})
\end{equation}
\begin{equation}\label{eq:pos_encode_odd}
PE_{pos,2i+1}=\textit{cos}(\frac{pos}{10000^{2i/D}})
\end{equation}
This positional encoding for each odd and even position is then added to the feature vector to allow the neural network some context into the order of inputs. There was no need to form a tokenised input embedding since we already have a distinct feature space described in our feature vector directly from the tabular sequences.

\begin{figure*}[t!]
  \centering
  \includegraphics[width=1\textwidth]{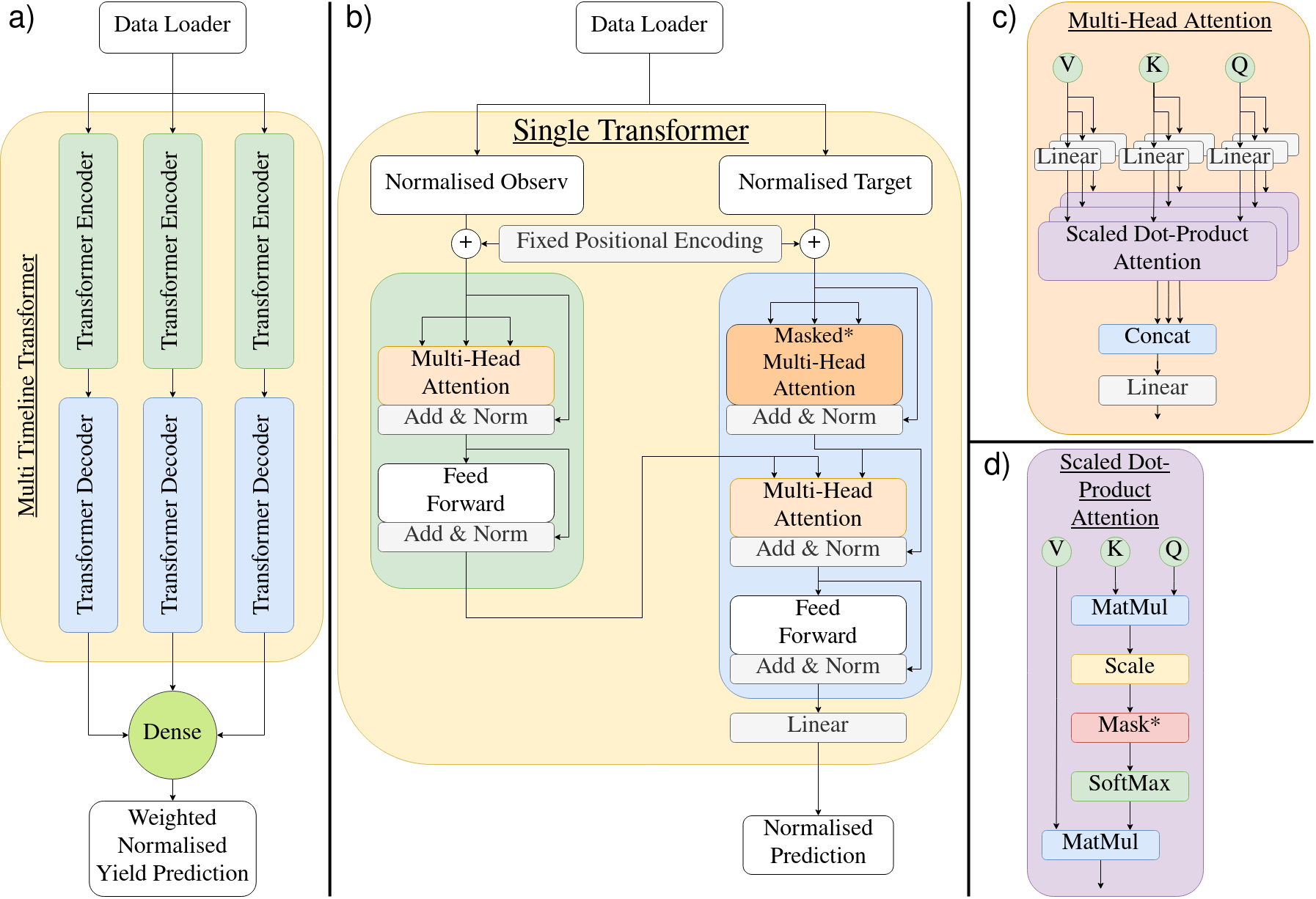}
  \caption{a) Mutil Timeline Transformer (MTT) architecture wherebye three single transformers that each process different data streams, are merge by a learned dense layer to weight their significance. b) A full single transformer architecture comprised of fixed positional encoding, encoder, decoder, and linear layers notably missing Softmax. c) Multi-head attention mechanism with query, key, and value matrices. This is a sub-components of transformer encoder and decoders with optional masks to maintain the temporal blindness when processing all the data simultaneously. d) Scaled dot-product attention showing the various matrix operations necessary to compute. this is a sub component of multi-head attention.}
  \label{fig:mtt}
\end{figure*}

An abbreviated form of the multi head attention depicted in Figure \ref{fig:mtt} (c) is Equation \ref{eq:multi_head_attention} along with the weight matrices.
\begin{gather}\label{eq:multi_head_attention}
\textit{Attention}(Q,K,V) = \textit{softmax}(\frac{QK^T}{\sqrt{d_k}})\\
W_i^Q \in \mathbb{R}^{D{\times}d_k}\\
W_i^K \in \mathbb{R}^{D{\times}d_k}\\
W_i^V \in \mathbb{R}^{D{\times}d_v}
\end{gather}

\subsubsection{Dense}
The dense layer is a simple linear layer with enough weights to form the weighted sum of the inputs and concatenate them into a singular value output in Equation \ref{eq:linear}
\begin{gather}\label{eq:linear}
\hat{y} = \sum{a_{t}W_{t}} + \sum{a_{n}W_{n}} + \sum{a_{f}W_{f}}
\end{gather}

Towards gathering data we employed our own data collection pipeline on our \anon{Riseholme} strawberry tabletop site, the respective yields of this site can be seen in Figure \ref{fig:7davg}. All the following data is streamed into MongoDB and accessed using aggregation pipelines to help speed up the transformation process.

\subsubsection{Weight Initialisation}

For weight initialisation we used the default pytorch Kaiming uniform initialisation as defined in Algorithm \ref{alg:kaiming} for leaky-ReLU (\autocite{relu, leakyrelu}).

\begin{algorithm}[h]
	\caption{Kaiming uniform weight initialisation using leaky-ReLU with the fan-in method. where $a$: (default 0 for ReLU, or -0.01 for leaky-ReLU) is the negative slope of the rectifier used after this layer. $W$: a randomised weight matrix with mean 0 and variance 1 (shape e.g $(64,32)$) $\textit{mode}$: is a flag which represents a different value for the $\textit{fan}$ whether the method being used is for feedforward or backpropagation (e.g if $\textit{mode}=\text{fanin}$ then $\textit{fan}=64$ else $\textit{fan}=32$ given previous example $W$ matrix).}\label{alg:kaiming}
	\begin{algorithmic}
		\Function{kaiming\_uniform\_weight\_init}{$a$, $W$, $\textit{d}$}
		\If{$\textit{mode} = \textit{fanin}$}
		    \State $\textit{fan} = dim(W, 0)$
		\Else
		    \State $\textit{fan} = dim(W, 1)$
		\EndIf
		\State $\textit{std} = \sqrt{\frac{2}{(1+a^2)\times\textit{fan}}}$
		\State \Return{$W\star\textit{std}$}
		\EndFunction
	\end{algorithmic}
\end{algorithm}

\subsubsection{Loss Function}

We chose to use the Mean Squared Error (MSE) as our loss function where $\text{MSE}=\frac{\sum_{i=0}^{N-1} (y-\hat{y})^2 }{N}$. This allows us to exponentially penalise large more errors than small errors on our continuous yield forecast. We in particular seek to reduce the networks tolerance for larger single errors as these would mean even if the total error was the same, being particularly peaked in one prediction would result in the growers having to import fruit that particular week. We would much rather be consistently out by a known amount than having almost perfect performance one week and then large errors the next.

As is commonly the case we use Adaptive moment (ADAM \autocite{kingma2014adam}) as our neural network optimiser as it is has been shown to be more performant than just first order or second order moments and is by and large the defacto standard. We calculated our first order moments $ m_t = \beta_1 * m_{t-1} + (1-\beta_1) * g_t $ $ \hat{m_t} = \frac{m_t}{1 – \beta_1^t} $ and second order moments.

\subsection{Models}

\begin{figure*}[h]
  \centering
  \includegraphics[width=1\textwidth]{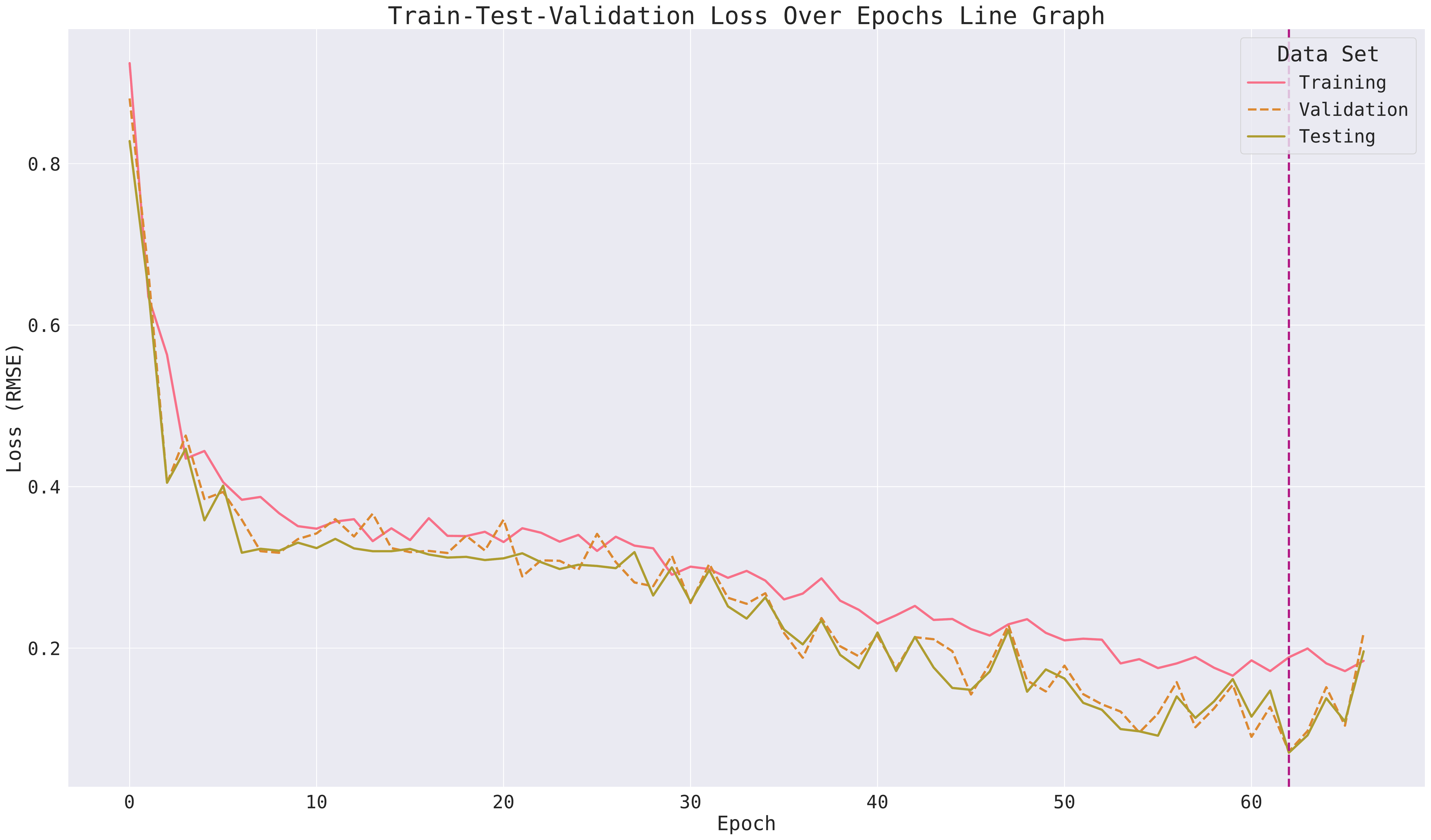}
  \caption{Three timeline transformer loss training, validation and testing sets, per epoch of training. Beyond 62 epochs (pink vertical line) validation and testing loss steeply increases again.}
  \label{fig:model}
\end{figure*}

We primarily focused on two different types of model. One holistic model that learned from all of the training rows using random subsets for training and validation (Figure \ref{fig:model}). Then we also attempted to create smaller weaker predictors as an ensemble only trained on a smaller set of the training data to each other as an ensemble to attain simple certainty metrics, which we deem would be invaluable towards building trust in the models and enabling re-investigation of uncertain scenarios. We split the training data used into 3 row sets of tabletop for each ensemble member. Each ensemble member is equivalent to the base MTT, including weight initialisation, loss function, and optimiser. Overall this means there was a one-row overlap between the first-second and second-third MTT. The results of our two current attempted approaches along with our past approaches and expected forecasting performance of growers and agronomists can be seen in Table \ref{tab:errors}.

\begin{table}[h]
    \centering
	\begin{tabular}{lc}
        Forecaster                                       & Expected Error        \\ 
        \hline
        Grower                                           & 25\%$\dagger$         \\
        Agronomist                                       & 17\%$\dagger$         \\
        Recurrent Neural Network (RNN)                   & 21\%                  \\
        Long-Short Term Memory network (LSTM)            & 38\%                  \\
        Gated Recurrent Network (GRU)                    & 16\%                  \\
        \textbf{Multi-Timeline Transformer (MTT)}        & \textbf{8}\%          \\
        Ensemble of MTT (average)                        & 27\%                  \\
        Ensemble of MTT (median)                         & 30\%                  \\
    \end{tabular}
	\caption{Expected errors by forecasting source. All models are from our previous work trialling different methods on the same dataset. \\
		$\dagger$ : These are estimates and  may not be representative of any grower or agronomist specifically but are instead ballpark figures for illustration based on our information from our industry partners.}
    \label{tab:errors}
\end{table}

\begin{figure*}[h]
  \centering
  \includegraphics[width=1\textwidth]{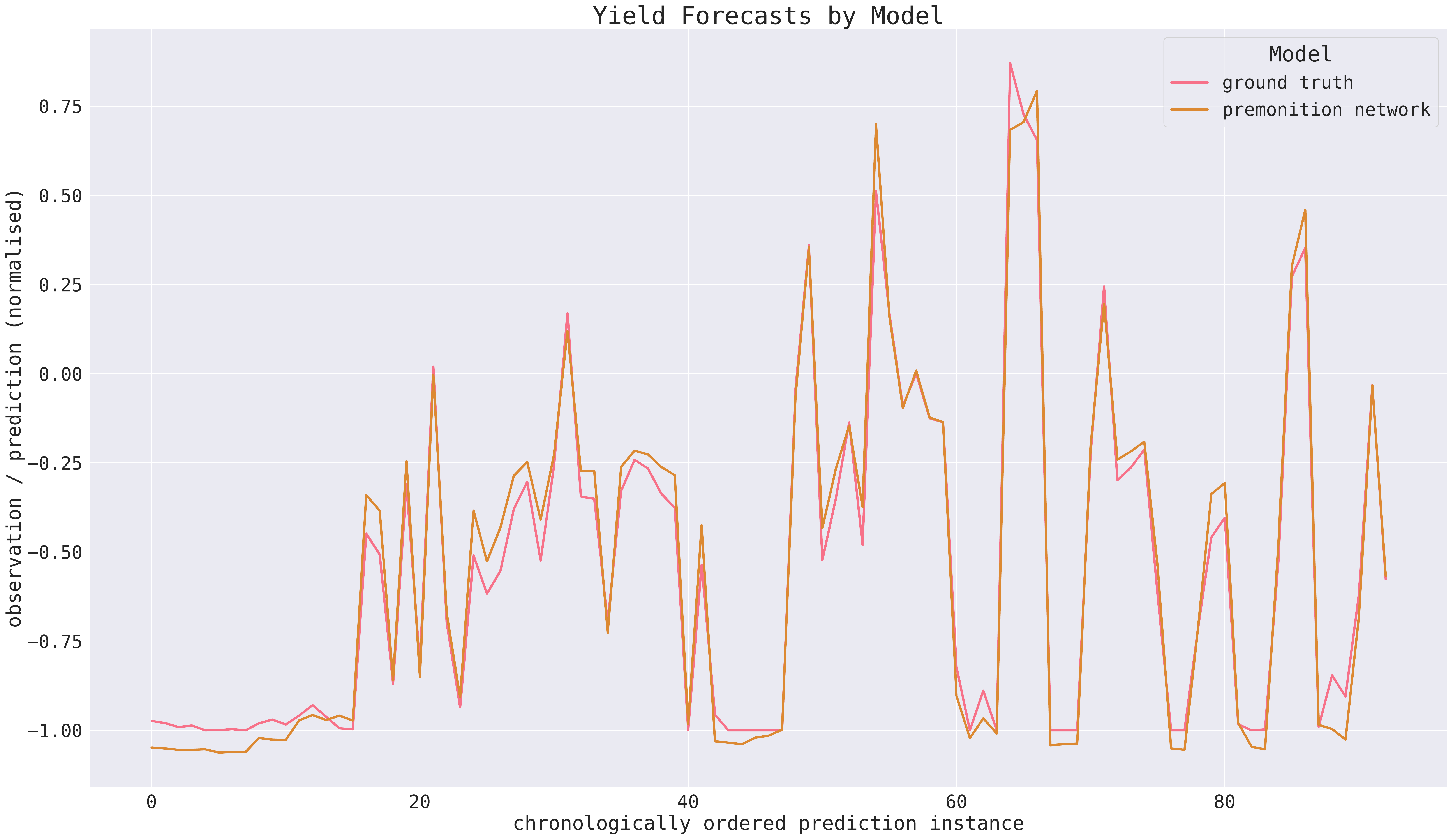}
  \caption{Ordered forecasts of single MTT compared to ground truth with a horizon of 3 weeks and a history of 12 weeks.}
  \label{fig:models}
\end{figure*}

\section{Discussion}\label{sec:discussion}
\subsection{Methodological Discussion}
Our strawberry dataset while covering 200 meters of strawberries is still limited. Commercial sites in comparison have hectares of such crops, meaning our 200 meters is not as representative of larger sites with more intra-crop variability. However as previously mentioned data availability is scarce making it practically very difficult to collect hectares of data, not least due to actual or perceived data sensitivity by the respective growers. In spite of this, while there may need to be some adjustments to account for more intra-crop variability of these larger sites our neural networks perform well given the data availability. While the sites are smaller and easier to learn, they also have fewer data to do so, which we believe to be a fair trade-off with no loss in difficulty between their larger sites and our smaller site.

Our MTT used an interval of 4 hours despite our data being synchronised over 15 minutes intervals. This was a tradeoff between data density (thus model complexity), and data availability. Since we only had a finite number of concrete outcomes that we observed we had to limit the complexity and weights of the model so that it could train its fewer weights with what limited data we had for concrete observations. In contrast, if we had used a data density of 15 minutes intervals we would have had to have significantly larger weight matrices being backpropagated from the limited number of observed yield values. If however, we found ourselves with large hectare scale datasets with many more observed outcomes we could tune the model to be more complex to leverage this data, to allow the model to understand much more complex relationships like the aforementioned expected intra-crop variability.

It may also be noted that we use a simple missing data imputation algorithm strategy namely forward-fill which involves filling missing values with the last known value. This was chosen as we mostly only incurred individual or relatively sparsely missing data. In larger sites one might expect to find entire regions that have some data unavailability for some time, meaning more advanced data-filling strategies may be necessary under such conditions. However, in our site, since the missing values were relatively sparse, the forward fill strategy is sufficient to allow us to leverage data in spite of any missing observations or features. The only notable exception is that of yield values. Since yield values were recorded sparsely a single missing value represents a much larger significance. Thus any such missing values are excluded entirely. Thankfully we had very few such missing values.

Due to the data scarcity we used fixed positional encoding as opposed to learned encoding. This means the gradients would not be shared with the learned positional encoding. This is sufficient since in the original transformer paper (\autocite{vaswani2017attention}) fixed positional encoding and learned positional encoding result in similar performance.

Finally, we chose to use a tri-transformer architecture merged using a dense fully connected layer. We did this to allow the neural network to train separate contextualising units for each potential timeline. This way we can easily conceptualise the timelines as follows. The pasts purpose is to have a broad view of the relationship between the features and the expected outcomes. This is important as we want to ensure the network has context for how yields are expected to outcome given past scenarios. The present serves to contextualise how this current specific season or crop is performing such that it can later be related to what has happened in the past. The future timeline / transformer is to add mitigations and adverse effects, such that high expected fluctuations can be considered at the merging layer.

\subsection{Results Discussion}

As can be seen in Table \ref{tab:errors} our primary MTT that can forecast three weeks ahead within 8\% RMSE is a large improvement over current capabilities as forecasts by agronomists tend to not only vary wildly from agronomist to agronomists (14 to 30\%), rely on specialist human presence, and are less accurate than our current model. However, a large caveat is that our model was created with intensive/high-quality environmental and yield data, on a small site compared to the typical industrial settings.

The results shown in Table \ref{tab:errors} and Figure \ref{fig:models} are a significant step forward in the prediction of strawberry yields, however, there are some weaknesses to our approach and the yield outcomes. Firstly our ensemble is significantly under-performing especially since a single predictor trained on the whole dataset beats the ensemble significantly. This is likely due to data, with almost three times the parameters we suspect that we require more training data to learn adequately, yet they receive $1/3$ of the total training data each. However, as time progresses and more data becomes available to us over more seasons we believe this ensemble will outperform the single MTT while enabling ensemble-based certainty estimation. Secondly and most difficult is the data itself. While we are fortunate to have access to our Risehome campus and the strawberry tabletop site, there is still a lack of data available for use. This relatively small site means we likely have not learned some of the more complex variances present on larger sites where the sensors immediate environment might be significantly different to another area on the growing site some distance away meaning the data in such scenarios might be significantly less representative of the conditions experienced by the strawberries.

Another issue with our dataset is the similarity between rows. Largely while there is inter-row variance there is still a risk of overfitting since even if the neural network cannot see row 1 for instance, it may be able to relate the yields of row 1 from previously trained/ known yields of row 2. We would have ideally liked to have split by time, and claimed one whole season as a completely separate testing set with none of those rows being trained on. However, due to the reality of strawberry seasonality and that there are only so many seasons with which it was possible to collect data, we had to split in such a way as to give the neural networks context for at least two seasons from start to finish. This is only necessary since the current methods of strawberry prediction in industry are largely based on the occurrences of the last season. As such we attempted to base our methods on existing techniques, and intuitively the performance of the strawberries last year will be related to the performance of the current season unless some large shift in methods between the seasons occurs.

One clear area of future work is to improve on data availability, whether that be collecting more data on a larger site and making that more easily available, or ordaining easily implementable data collection methods that can scale well, and are private such that data can be shared with little concern of sensitivity.

Another area of future work is to implement certainty metrics that do not require the use of ensembles so that we can keep the parameters down, and reduce the necessary data to train more complex models.
\ifx\blind\undefined
We also seek to make transformers that are abelian compatible such that we can use some of our prior fully homomorphically encrypted (FHE \autocite{gentrysFirstFhe}) deep learning methods with these currently incompatible but performant transformers \autocite{onoufriou2021fully, FHEoNNG}.
\else
\fi

Lastly, we seek to find ways in which to make our data available for wider use, currently that is not possible due to contractual constraints which were necessary to enable us to collect this data with industrial varieties in the first instance. However, we seek to remedy this in future.

\section{Conclusion}\label{sec:conclusion}
Neural networks are highly performant at predicting strawberry yields, here we attain 8\% RMSE with a three-headed transformer ingesting three distinct timelines/ streams. The past, the present, and the future. We collected our own data to fuel these timelines due to the distinct unavailability or impracticality of currently available strawberry yield data. Our multi-timeline approach achieves better performance than humans by a significant margin. However, we also found that there is still an ongoing battle with data availability which still restricts us from certain model architectures due to insufficient data with which to train them.


%


\ifCLASSOPTIONcompsoc
  \section*{Acknowledgments}
\else
  \section*{Acknowledgment}
\fi
\ifx\blind\undefined
    \textbf{Funding statement}: This research was supported by the Biotechnology and Biological Sciences Research Council (BBSRC) studentship [grant numbers \href{https://gtr.ukri.org/projects?ref=studentship-2155898}{2155898}, \href{https://gtr.ukri.org/projects?ref=BB/S507453/1}{BB/S507453/1}]. The authors would also like to thank the \href{https://lcas.lincoln.ac.uk/wp/}{Lincoln Centre for Autonomous Systems (L-CAS)} for their help and support with the autonomous robotic \href{https://sagarobotics.com/}{Thorvald} systems, and the availability of tools and resources such as the Riseholme strawberry tabletop itself.
\else
    \textbf{Funding statement}: Anonymised
\fi
\\

\textbf{Author contributions:} Conceptualisation, \anon{G.O.} and \anon{G.L.}; methodology, \anon{G.O.}; software, \anon{G.O.}; validation, \anon{G.O.}; formal analysis, \anon{G.O.}; investigation, \anon{G.O.}; resources, \anon{G.O.}, \anon{G.L.} and \anon{M.H.}; data curation, \anon{G.O.}; writing—original draft preparation, \anon{G.O.} and \anon{G.L.}; writing—review and editing, \anon{G.O.} and \anon{G.L.}; visualisation, \anon{G.O.}; supervision, \anon{G.L.}; project administration, \anon{G.L.} and \anon{M.H.}; funding acquisition, \anon{G.L.} and \anon{M.H.} All authors have read and agreed to the published version of the manuscript.\\

\textbf{Conflict of interest}: The authors declare no conflict of interest.

\ifCLASSOPTIONcaptionsoff
  \newpage
\fi



%


\printbibliography

%









\end{document}